\def\year{2020}\relax
\documentclass[letterpaper]{article} 
\usepackage{aaai20}  
\usepackage{times}  
\usepackage{helvet} 
\usepackage{courier}  
\usepackage[hyphens]{url}  
\usepackage{graphicx} 
\usepackage{graphicx}  
\usepackage{color}
\usepackage{soul}
\usepackage{amssymb}
\usepackage{amssymb}
\usepackage{ulem}
\usepackage{array}

\title{Explainable AI for System Failures: Generating Explanations that Improve Human Assistance in Fault Recovery} 
 
\author{Devleena Das, Siddhartha Banerjee, Sonia Chernova\\
Georgia Institute of Technology, Atlanta, Georgia\\
Institute for Robotics and Intelligent Machines\\
\{ddas41,siddhartha.banerjee, chernova\}@gatech.edu 
}
 
\begin{document}

\maketitle
\begin{abstract}
With the growing capabilities of intelligent systems, the integration of artificial intelligence (AI) and robots in everyday life is increasing. However, when interacting in such complex human environments, the failure of intelligent systems, such as robots, can be inevitable, requiring recovery assistance from users. In this work, we develop automated, \textit{natural language} explanations for failures encountered during an AI agents' plan execution. These explanations are developed with a focus of helping \textit{non-expert} users understand different point of failures to better provide recovery assistance. Specifically, we introduce a context-based information type for explanations that can both help non-expert users understand the underlying cause of a system failure, and select proper failure recoveries. Additionally, we extend an existing sequence-to-sequence methodology to automatically generate our context-based explanations. By doing so, we are able develop a model that can generalize context-based explanations over both different failure types and failure scenarios.


\end{abstract}

\section{Introduction}
In homes, hospitals, and manufacturing plants, robots are increasingly being tested for deployment alongside non-roboticists to perform goal-directed tasks, such as folding laundry \cite{yang2016repeatable}, delivering laboratory specimens \cite{bloss2011mobile,hu2011advanced}, and moving inventory goods \cite{hagele2016industrial,lawton2016collaborative}. 
When interacting in such complex human environments, robot failures are inevitable and assistance in failure recovery can be necessary \cite{bauer2008human}. An example of such a scenario is that of a consumer interacting with technology in their own home, such as determining why a robot tasked with retrieving a beverage is stopped in the middle of the kitchen, or a scenario where a production line worker wonders why a robot who was picking up boxes from a conveyor belt moments ago, suddenly stopped. Prior work in the Explainable Planning (XAIP) community has explored closely related problems, such as establishing methods for explaining an agent's chosen plan for a particular task, and explaining unsolvable plans to end-users \cite{chakraborti2020emerging}. However, providing justifications for points of failures that occur \textit{during} an agent's plan execution has not yet been studied. 

In this work, we aim to expand upon the existing set of explanations available in the XAIP community. We propose an additional type of explanation called \textit{error explanations}, in the context of sequential-decision making and planning. These error explanations focus on explaining failures that may occur while executing a chosen plan. We seek to develop automated, \textit{natural language} error explanations that can explain encountered failures in a manner that is understandable by non-expert users. The goal of these explanations is to not only help non-expert users understand the system's point of failure, but also help them determine an appropriate solution required to resume normal operation of a task. Specifically, our core research questions are:
\begin{itemize}
    \item What type of information constitutes a meaningful explanation of an agent's failure that can aid in a \textit{non-expert's} ability to understand the cause of a failure, and provide accurate fault recovery assistance? 
 
    \item How can we develop a model that can automatically generate natural language explanations so that these explanations can be generalized across varying failure scenarios and failure types?
\end{itemize}
Through these fundamental questions, we i) introduce a context-based information type that explanations should include to effectively help users understand the fault diagnoses and in turn provide accurate recovery assistance, and ii) adapt an existing sequence-to-sequence methodology from \cite{ehsan2019automated} to generate automated explanations that can generalize over varying failure types and scenarios.

We validate our approach through a user study, comparing two different types of explanations, action-based and context-based, applied to a pick-and-place robot manipulation task. Through this user study, we measure non-experts' accuracy in understanding the provided fault diagnoses and accuracy in identifying correct recovery solutions. We also measure users' self-reported confidence and difficulty scores for each decision. We observe that context-based explanations significantly improves users' recovery selection over both the baseline and action-based explanations. Additionally, self-reported ratings show that the presence of any explanations allows for higher perceived confidence and lower difficulty scores than having no explanations. Furthermore, the confusion matrix of our automated explanation generating model shows that our model can generalize over different failure scenarios with a 89.7\% overall accuracy. 


\section{Related Works}
In prior work, the XAI community has primarily focused on developing interpretability methodologies for expert users familiar with the domain of AI or ML \cite{adadi2018peeking,ribeiro2016should}. Many of these approaches have focused on model-agnostic implementations, designed to increase understanding of deep learning (DL) outputs for classification-based tasks by leveraging inherently interpetable models, such as, decision trees \cite{zhang2019interpreting}, or visual attributes, such as, heatmaps \cite{selvaraju2017grad}. While these approaches are applied to more complex models, the complexity of such classification tasks do not include the complexity of sequential decision-making, long-term interactions, or changing environments \cite{chakraborti2020emerging}.

Current work in XAIP aim to address the need for interpretable explanations for complex planning problems which expand beyond single-classification tasks. In a recent survey paper, \cite{chakraborti2020emerging} highlight some of the key components of plan explanations studied by the community: contrastive question-answering, explaining unsolvable plans, and achieving explicable justifications for a chosen plan. In the realm of answering contrastive questions, \cite{krarup2019model} describe a framework to transfer domain-independent user questions into constraints that can be added to a planning model, while \cite{hoffmann2019explainable} describe how to utilize common properties within a set of correct plans as an explanation for unmet properties in incorrect plans. In order to explain unsolvable plans, \cite{sreedharan2019can} abstract the unsolvable plan into a simpler example through which explanations can be formulated. Additionally, \cite{zhang2017plan} describe the need for explanations to be ``explicable" by end-users. The authors of this work implement explicability by using conditional random fields (CRFs) to model humans' labelling schemes to agent plans, and use such model to develop explicable explanations for new plans. Additionally, to minimize the constraints on an agent's plan, \cite{chakraborti2019explicability,chakraborti2017plan} describes a particular process of achieving explicability, known as model reconciliation. The authors produce explanations by considering the difference between an agent's and end user's mental model. In all these cases, a chosen plan, or lack thereof, is explained. In our work, instead of explaining a particular plan, we aim to explain possible faults within a plan that consequently halt its execution.

Outside the scope of XAIP and in the context of reinforcement learning systems, \cite{ehsan2018rationalization,ehsan2019automated} also describe the need for humanly understandable explanations. The authors coined the usage of \textit{rationales} as a way of generating explanations in language that is understandable by everyday people. They developed an automated rationale generating system, studied within the context of the game Frogger, that can translate game state representations into humanly understandable explanations. However these explanations are generated within the domain of discrete-action state space and not continuous-action state space which are commonly found in sequential decision-making, planning problems.

Furthermore, within the realm of fault recovery in robotics, \cite{knepper2015recovering} studies how robots can utilize natural language to generate assistance requests during a point of error. Their natural language framework is trained to generate assistance requests with accurate multi-object disambiguation ('table leg under the table' vs. 'table leg near the table') in efforts  shorten idle times during assembly. Instead of focusing on object disambiguation or asking for a specific assistance, we utilize natural language to generate explanations that can explain a robot's failure in a manner that allow non-expert users to deduce a plausible recovery assistance.


\renewcommand{\arraystretch}{1.5}

\begin{center}
\begin{table*}[h!]
\begin{tabular}{|>{\centering\arraybackslash}m{2.0cm}|>{\centering\arraybackslash}m{2.2cm}|m{4.4cm}|m{7.3cm}|}
\hline
\multicolumn{1}{|c|}{Failure Type} &\multicolumn{1}{c|}{Scenario} &\multicolumn{1}{c|}{Action-Based} & \multicolumn{1}{c|}{Context-Based}\\
 \hline
 motion-planning & obj. too far away& Could not move its arm to the desired object& 
Could not move its arm to the desired object because the desired object is too far away 
\\ \hline
 motion-planning& obj. close to other objs. &
Could not move its arm to the desired object
&
Could not move its arm to the desired object because the desired object is too close to other objects
 \\ \hline
detection & obj. not present& Could not detect the desired object& 
Could not detect the desired object because the desired object is not present where the robot is looking
 \\ \hline
 detection   & obj. occluded & Could not detect the object& 
Could not detect the desired object because the desired object is occluded 
 \\ \hline
 navigation& mis-localization & Could not navigate to the desired object& 
Could not navigate to the desired object because the robot is lost
 \\ \hline
  navigation & controller &Could not navigate to the desired object& 
Could not navigate to the desired object because the robot’s motors are malfunctioning
 \\ 
\hline
\end{tabular}
\caption{Example explanations for each failure type and failure scenario that are provided to the AB and CB study conditions.}
\label{tab:Exps}
\end{table*}
\end{center}
\section{Problem Definition}

Building on the definition presented by \cite{chakraborti2020emerging}, we define a planning problem $\Pi$ in terms of a transition function $\delta_\Pi : A \times S \rightarrow S \times \mathbb{R}$, where $A$ is the set of actions available to the agent, $S$ is the set of states it can be in, and the real number denotes the cost of making the transition. A planning algorithm $\mathbb{A}$ solves $\Pi$ subject to a desired property $
\tau$ to produce a plan or policy $\pi$, i.e. $\mathbb{A} : \Pi \times \tau  \mapsto \pi$. Here, $\tau$ may represent different properties such as soundness, optimality, etc.  The solution to this problem is defined as a \textit{plan} $\pi = \langle a_1, a_2, ..., a_n \rangle, a_i \in A$, which transforms the current state $I \in S$ of the agent to its goal $G \in S$, i.e. $\delta_\Pi (\pi, I) = \langle G, \Sigma_{a_i \in \pi} c_i \rangle$. The second term in the output denotes the plan cost $c(\pi)$.

In this context, we argue that there are (at least) two categories of explanations that are useful to a user.  The first was included in the survey by \cite{chakraborti2020emerging}, and the second we introduce here:
\begin{itemize}
    \item \textbf{$\mathcal{E}_\pi$:} This explanation serves to justify to a human user that solution $\pi$ satisfies property $\tau$ for a given planning problem $\Pi$.  For example, the user may ask ``Why $\pi$ and not $\pi'$?''.  In response to this question, $\mathcal{E}_\pi$ must enable the user to compute $\mathbb{A} : \Pi \times \tau  \mapsto \pi$ and verify that either $\mathbb{A} : \Pi \times \tau  \not \mapsto \pi'$, or that $\mathbb{A} : \Pi \times \tau  \mapsto \pi'$ but $\pi \equiv \pi'$ or $\pi > \pi'$ with respect to some criteria.  $\mathcal{E}_\pi$ applies to the plan solution as a whole and can be elicited at any time.  Approaches that address $\mathcal{E}_\pi$ are listed in the Related Works section.  
    
    \item \textbf{$\mathcal{E}_{err}$:} This explanation applies in the event that an unexpected failure state $f \in\mathcal{F}$, triggered by a failed action in $\langle a_1, a_2, ..., a_n \rangle$, halts the execution of $\pi$.  For example, the user may ask ``The robot is at the table, but why did it not pick up my beverage?" In response to this question, $\mathcal{E}_{err}$ must allow the user to understand the cause of error in order to help the system recover.
    
\end{itemize}

In this work, we address the second variant of explanations, $\mathcal{E}_{err}$. We assume that both the algorithm $\mathbb{A}$ and the policy $\pi$ are sound, and that the cause of error is triggered by a failure state $f \in \mathcal{F}$ from which it cannot recover without user assistance.  Our objective is to find $\mathcal{E}_{err}$ such that the user correctly understands the cause failure, and can help the agent recover from an error.  We introduce a set of \textit{information types} $\Lambda$ that evaluate varying characteristics of an explanation {$\mathcal{E}_{err}$} in order to find a meaningful $\lambda^{'} \in \Lambda$ for non-expert users. To generalize and automate an explanation {$\mathcal{E}_{err}$} for different failure scenarios, we take inspiration from \cite{ehsan2019automated}'s work to translate the state of the agent, $S$, into natural language explanations that fit $\lambda^{'}$.

\section{ Information Types for $\mathcal{E}_{err}$ }

The first question we have to answer is: given an error while executing $\pi$, what format should explanation $\mathcal{E}_{err}$ take?  

\cite{ehsan2019automated}'s work establishes that explanations for everyday users should take the form of rationales that justify a reasoning in layperson's terms while being representative of a particular scenario, as opposed to revealing the true decision making process of an agent. Thus, to provide an effective and meaningful {$\mathcal{E}_{err}$} to non-experts, we first evaluate a set of information types $\Lambda$ to find the best information type $\lambda^{'}$ that {$\mathcal{E}_{err}$} should encompass. For this, we conducted a three-way between-subjects user study where participants were asked to identify and suggest fixes to a set of failure states $\mathcal{F}$ that a robot encounters while performing $\pi$. In this study design, $\Lambda$ consists of the following three study conditions that differ the information type of {$\mathcal{E}_{err}$}:
\begin{itemize}
    \item \textbf{None (Baseline)}: Participants receives no explanations on the cause of error.
    \item \textbf{Action-Based (AB)}: Participants receive {$\mathcal{E}_{err}$} that use the failed action as the cause of error, seen in Table \ref{tab:Exps}. 
    \item \textbf{Context-Based (CB)}: Participants receive {$\mathcal{E}_{err}$} that use the failed action as well as a contextualized reasoning deduced from the environment as the cause of error, seen in Table \ref{tab:Exps}.
\end{itemize}
To validate which type of $\mathcal{E}_{err}$ is most meaningful, we conducted an experiment using simulated robot errors and scripted explanations.  In the subsections below, we present our experimental framework, the study design, and the results. This evaluation serves to inform the $\lambda^{'}$ that will of focus when developing an automated generation of $\mathcal{E}_{err}$.


\subsection{Metrics}
We use the following metrics to evaluate the effectiveness of a particular $\mathcal{E}_{err}$:

\begin{itemize}
    \item \textbf{Solution Percentage (Sol\%)}: measures how accurately participants select solutions to recover the encountered failure. The average solution percentage is calculated as:
    \begin{equation}
   {Sol\%}= \frac{correctSolution}{correctSolution + incorrectSolution}
   \end{equation}
    \item \textbf{Action Identification Percentage (\textit{AId \%)}}: measures how accurately participants identify the action on which plan $\pi$ fails. The average action identification percentage is calculated as:
    \begin{equation}
    {AId\%}= \frac{correctAction}{correctAction + incorrectAction}
    \end{equation}
    \item \textbf{Action Confidence (\textit{ActConf})}: measures self-reported confidence in determining a failed action in $\pi$. Action confidence is measured using a 5-Point Likert Scale rating based on the question "How confident are you in determining the failed action?" (1= Not Confident, 5=Very Confident).
    \item \textbf{Difficulty Rating (\textit{DiffRate})}: measures self-reported difficulty in determining a plausible solution to the encountered failure. Difficulty rating is measured using a 5-Point Likert Scale rating based on the question "How difficult was it to determine a solution to the encountered failure?" (1 = Not Difficult, 5=Very Difficult).

\end{itemize}

We hypothesize that the presence of AB or CB explanations will lead to high action identification scores \textit{(AId\%)}, compared to no explanations. However, we believe that in determining a plausible solution to an encountered failure \textit{(Sol\%)}, those with CB explanations will perform better due to the additional contextual reasoning they are provided. We also believe that both confidence and difficulty ratings will correlate highly with respect to each conditions' action identification and solution percentages. That is, CB and AB will have comparable confidence \textit{(ActConf)}, but CB participants will have lower perceived difficulty \textit{(DiffRate)} than AB participants.

\section{Experimental Setup}
\begin{figure}
\centering
  \includegraphics[width=0.9\columnwidth]{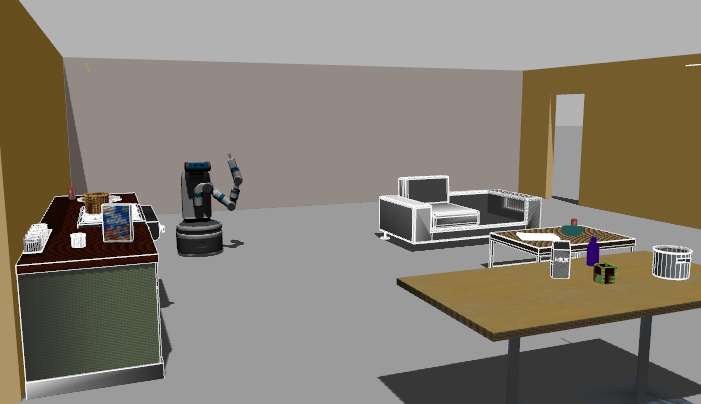}
  \caption{Home and kitchen environment in which Fetch performs a pick-and-place manipulation task.}~\label{fig:GazEnv}
\end{figure}

Our experimental setup uses a Gazebo simulation of a Fetch robot in a household setting performing a pick-and-place task (Figure \ref{fig:GazEnv}).  Similar to prior work in robotics  \cite{banerjee2020taking}, the robot's action set $A = \{move, segment, detect, findgrasp, grasp, lift, place\}$, where $move$ navigates the robot to a specified location, $segment$ is a perception action performed by the robot to identify which pixels in its sensory space correspond to objects, $detect$ performs object detection to obtain a label for a given object, $findgrasp$ executes grasp sampling to identify possible grasp poses for the gripper, $grasp$ moves the robot arm into a grasp pose and closes the gripper, $lift$ raises the arm, and $place$ places a held object at a specified location.



The robot's state at each time step $t$ is defined as $s_{t} \in S$, where $S = S_{e} \cup S_{l} \cup S_{i} \cup S_{k}$ describe the entities in the environment, the location of each entity, the agent's internal states and the task states, respectively. $S_{e}$ denotes the set of names for all entities in the environment, and does not change during the execution of $\pi$. We additionally define $S_{o} \subset S_{e}$ as the specific objects of interest to our agent, and $S_{p} \subset S_{e}$ as the semantic places of interest to the agent. $S_{o}$ is defined as: \textit{milk, coke can, ice cream, bottle, cup}, and $S_{p}$ is defined as: \textit{dining table, left kitchen counter}. $s_{l}(t) \in S_{l}$ is a vector of $\langle x , y, z \rangle$ locations of each entity $s_{e} \in S_{e}$ at a given time step $t$. $s_{i}(t) \in S_{i}$ is defined by three tuples $\langle x_{avel},y_{avel},z_{avel}\rangle$, $\langle x_{lvel},y_{lvel},z_{lvel}\rangle$ ,$\langle x_{pos},y_{pos},z_{pos} \rangle$ that describe the angular velocity, linear velocity and position of the agent at $t$. Finally, $S_{k} = \{k_{grasp},k_{findgrasp},k_{move},k_{pick},k_{detect},k_{seg}\}$ where $s_{k}(t) \in S_{k}$ describes the status of each $a \in A$ at $t$, and whether each action is: active ($0$), completed ($1$) or errored (-$1$). Therefore, at all time steps, the number of elements in $s_{k}(t)$ is equal to the number of actions in $A$.
\begin{center}
\begin{table}
\begin{tabular}{ |p{2.3cm}|p{1.8cm}|p{3.2cm}|}
\hline
 Failure Type & Failed action & Failure Scenario\\
\hline
 motion planning  & grasp    & obj. too far away \\
 motion planning& grasp & obj. close to other objs.  \\
 detection &detect& obj. not present \\
 detection   &detect & obj. occluded \\
 navigation& segment& mis-localization \\
 navigation & move  & controller \\
\hline
\end{tabular}
\caption{Description of the associated failure types, and the robot's failed action for each specific failure scenario. }
\label{tab:Failures}
\end{table}
\end{center}

\subsection{Simulating Failures}
In this work, the agent's initial state is defined as  $s_{0} = \{\langle 0,0,0\rangle,\langle 0,0,0\rangle,\{null\}\}$, where the position tuple and the velocity tuples are set to zero, and the tasks states $s_{k}(0)$ are not defined. The agent's final state is defined as $s_{T} = \{\langle x_{T},y_{T}, z_{T}\rangle, \langle 0,0,0\rangle, \{1,1, ..., 1\}\}$, where the position tuple is set to the goal location, the velocity tuple is zero and the each task state in $s_{k}(T)$ is 1. With these assumptions, we define a failure $f$ in plan $\pi$ when any task state in $s_{k}$ has a value -1.


Previous work in fault diagnosis has summarized possible categories of faults that may occur in a given $\pi$. We specifically focus on \textit{Component Faults} and \textit{Contextual Faults}. While the former describe hardware or software module failures, the latter describe failures caused by changes in the environment \cite{banerjee2019fault}. Table~\ref{tab:Failures} lists the type of failures $F_{t}$, the scenarios $F_{s}$ that can cause each type of failure, and the action on which $\pi$ fails. For the purposes of our experimentation, we simulate the navigation errors as \textit{Component Faults} caused by an error in the navigation software module, and the motion-planning and detection errors as \textit{Contextual Faults}. We define two failure scenarios per failure type, reflecting on the fact that a given failure type may have multiple causes. We denote $\mathcal{F} = size(S_{o}) \times size(F_{s})$ to be the set of all possible failure states, where $S_{o}$ is the objects of interest and $F_{s}$ are the failure scenarios.

\subsection{Presenting Explanations to Users}
For each study condition, participants were shown both failure scenarios from $\mathcal{F}$ as well as successful executions of $\pi$ for the given task objective. Participants watched three videos of Fetch successfully executing $\pi$ with randomly selected objects from $S_{o}$. The motivation for showing successful iterations of $\pi$ was to show participants that the plan $\pi$ itself was complete and executable. The remainder of the study consisted of identifying failure scenarios. Participants watched twelve videos, corresponding to twelve randomly chosen failure scenarios from $\mathcal{F}$. After each video, participants were presented with questions asking them to identify: the action $a$ that prompted the failure in $\pi$, a solution to the encountered failure, and their perceived difficulty of the questions and perceived confidence of their answers. 

\subsection{Participants}
We recruited 45 individuals from Amazon's Mechanical Turk, who were split into the three experimental groups. Our participants included 27 male and 18 female, who were all 25 or older. Specifically, 24 between  25- 34 years, 7 between 35-44 years, 8 between 45-54 years, and 6 who were 55 years or older. The task took participants approximately 20-25 minutes on average and they were compensated with \$2.50. 


\begin{figure}[t!]
\centering
  \includegraphics[width=0.9\columnwidth]{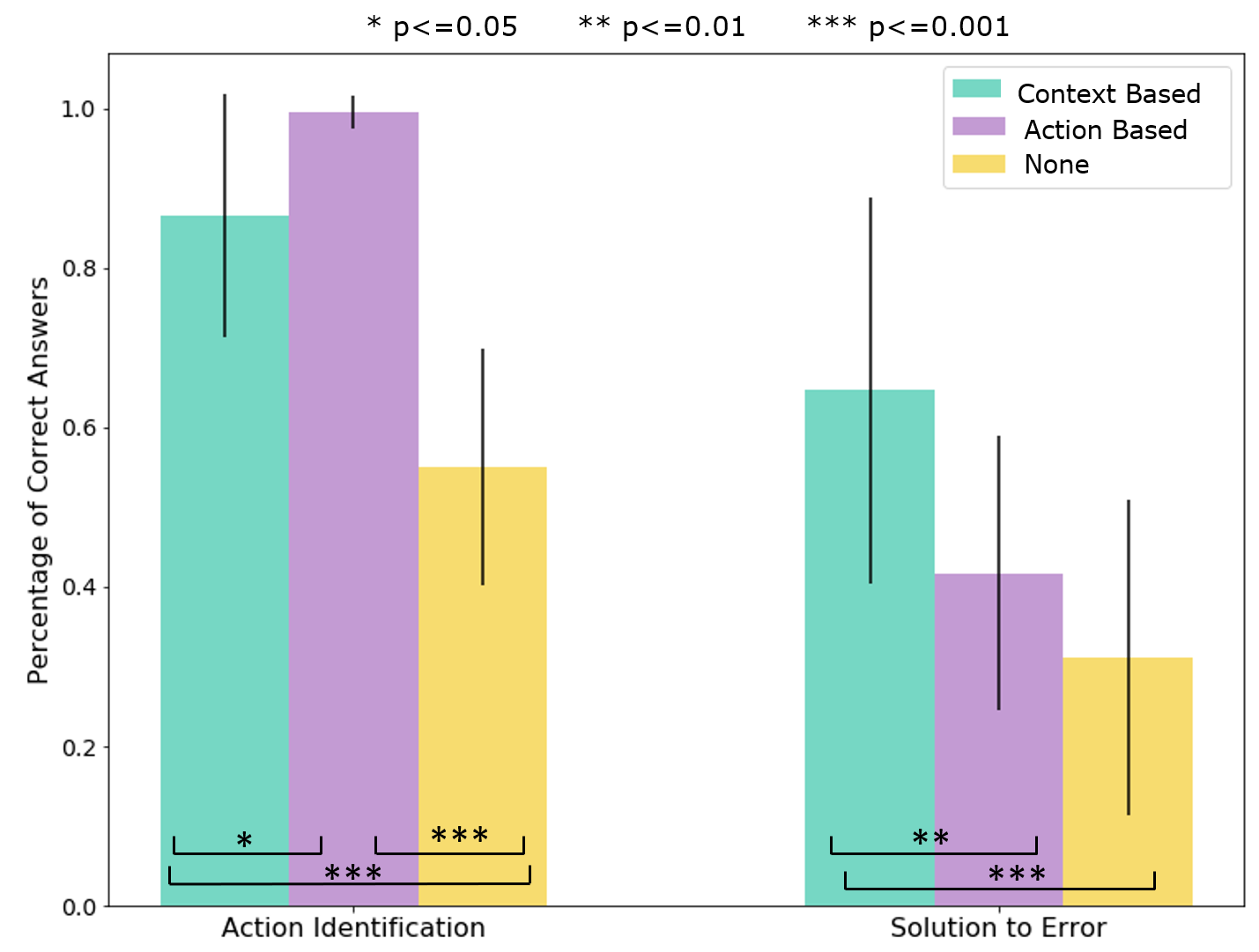}
  \caption{Average \textit{AId\%}  and \textit{Sol\%} across the conditions, where AB and CB participants were presented with an {$\mathcal{E}_{err}$}.}~\label{fig:CorrAnsP1}
\end{figure}
\begin{figure}[t!]
\centering
  \includegraphics[width=0.9\columnwidth]{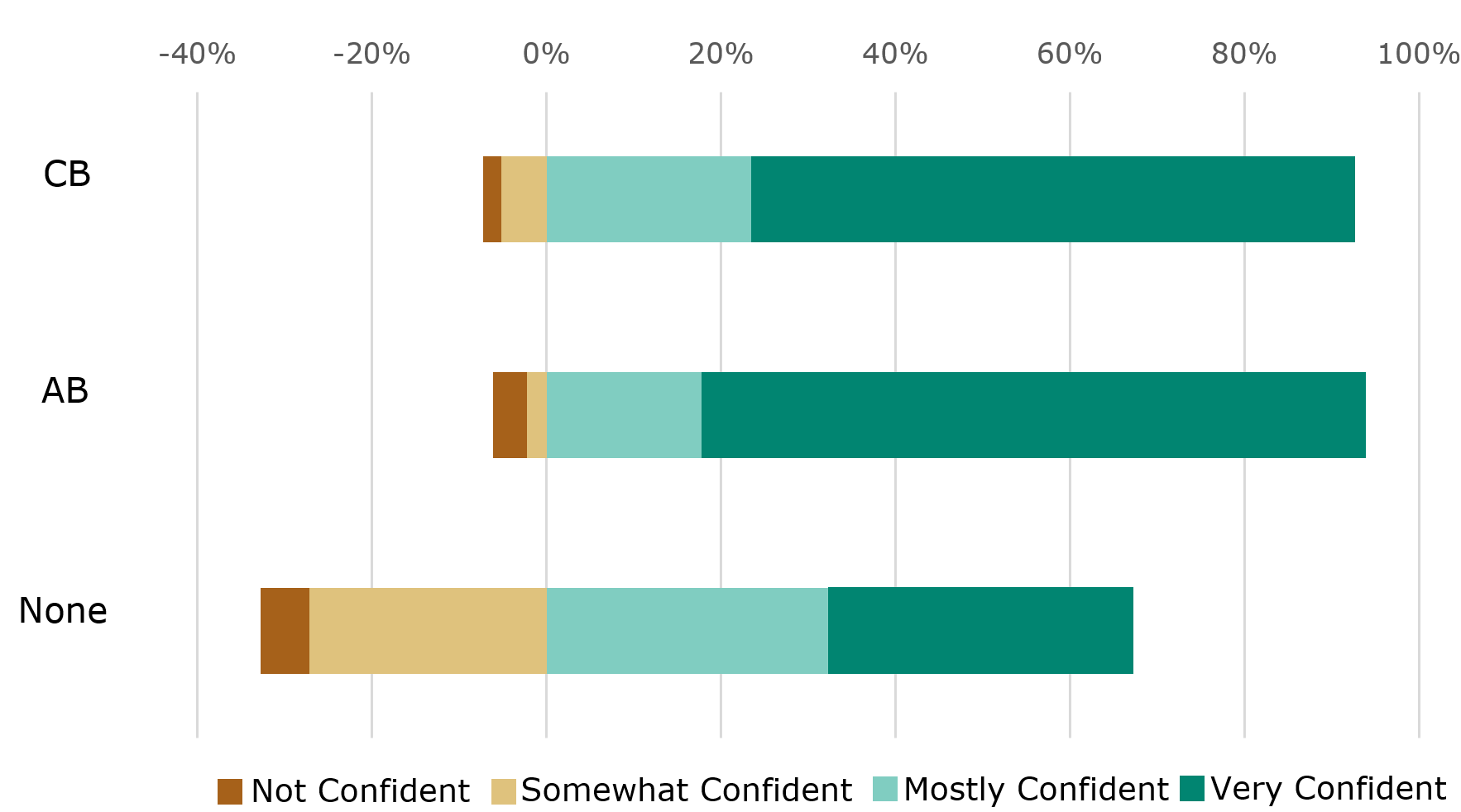}
  \caption{Participant's \textit{ActConf} across the conditions where AB and CB participants were presented with an {$\mathcal{E}_{err}$}.}~\label{fig:Likert-Conf}
\end{figure}
\begin{figure}[t!]
\centering
  \includegraphics[width=0.9\columnwidth]{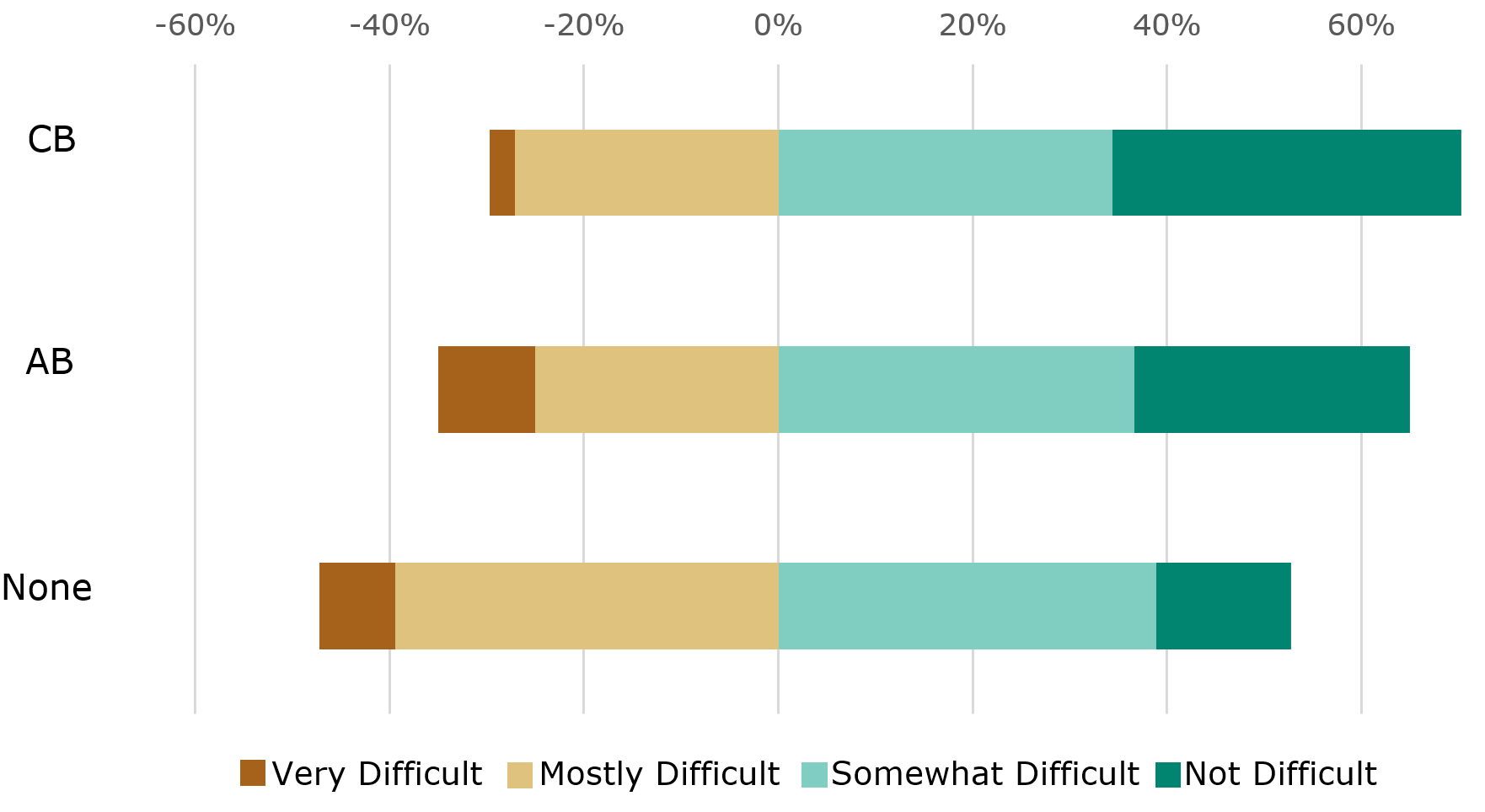}
  \caption{Participant's \textit{DiffRate} across the conditions where AB and CB participants were presented with an {$\mathcal{E}_{err}$}.}~\label{fig:Likert-Diff}
\end{figure}
\begin{center}
\begin{table}[t!]
\begin{tabular}{ |>{\centering\arraybackslash}m{2.4cm}||>{\centering\arraybackslash}m{2.3cm}|>{\centering\arraybackslash}m{2.3cm}|}
\hline
 \multicolumn{1}{|c||}{Conditions}& \multicolumn{1}{c|}{\textit{ActConf}} & \multicolumn{1}{c|}{\textit{DiffRate}}\\
 \hline
 CB vs. AB   & NS & NS\\
 CB vs. None & $p \le 0.001$ & $p \le0.001$\\
 AB vs None & $p \le 0.001 $& NS\\

\hline
\end{tabular}
\caption{ Mann-Whitney U test significance values for the \textit{ActConf} and \textit{DiffRate} metrics with a Bonferroni p-value correction. }
\label{tab:Signif}
\end{table}
\end{center}
\section{Experimental Results}
Since the participants' assessment data followed a normal distribution, we used ANOVA with a Tukey HSD post-hoc test to evaluate statistical significance for the  \textit{AId\%} and \textit{Sol\%}  metrics. To evaluate the statistical significance for the self-reported rating metrics, \textit{ActConf} and \textit{DiffRate}, we used  Kruskal-Wallis with a Mann-Whitney U post-hoc test and a Bonferroni correction.

Figure \ref{fig:CorrAnsP1} presents the average percentage of correctly identifying the failed action  (\textit{AId\%}), and correctly identifying a solution to the encountered failure (\textit{Sol\%}) for each study condition. We observe a significant difference in performance between the baseline (None) condition and the other conditions AB and CB. In other words, the presence of \textit{any} explanation {$\mathcal{E}_{err}$} helped participants better understand the failed action in $\pi$ and deduce possible solutions to errors  than those who were provided with no explanations. Additionally, we see that the inclusion of environmental context within an explanation (CB) significantly increased the accuracy of solutions to errors than explanations that only described the failed action (AB) in the plan $\pi$. This supports the idea that CB explanations help participants better understand the underlying cause of \textit{why} an error has occurred and therefore how to provide recovery assistance, as opposed to only knowing \textit{what} action caused the error within the system.

In Figure \ref{fig:Likert-Conf} and Figure \ref{fig:Likert-Diff}, we observe the self-reported ratings of how confident participants were in discerning the failed action, \textit{ActConf}, and how difficult it was to know the correct solution to a failure, \textit{DiffRate}. The Likert scale data shows that participants who were given an explanation (CB or AB), were more likely to rate of `Very Confident' and `Not Difficult', compared to the those who received no explanations (None). We also observe that AB and None participants had a similar number of `Very Difficult' ratings compared to CB participants, supporting that in the context of deducing a solution, AB explanations were not significantly more helpful than having no explanations (None). Our statistical analyses in Table~\ref{tab:Signif} support these conclusions, showing that \textit{any} explanation significantly improved participants' \textit{ActConf}, but only CB explanations were able to significantly improve participants' \textit{DiffRate}.

\begin{figure}[t!]
\centering
  \includegraphics[width=0.9\columnwidth]{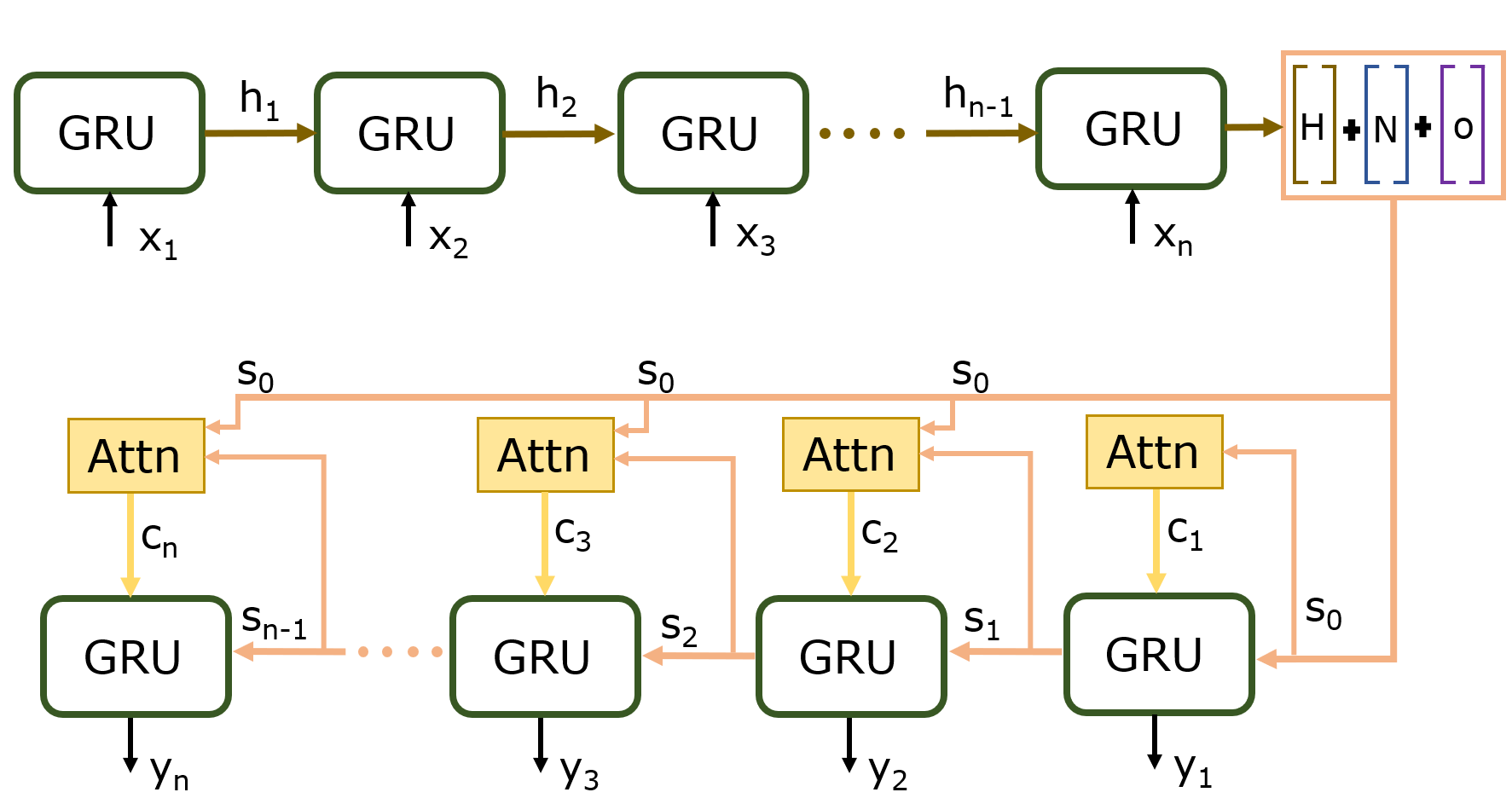}
  \caption{Our sequence-to-sequence model architecture for generating automated explanations.}~\label{fig:model-arch}
\end{figure}
\section{Automated Explanation Generation}
Our evaluations from above show that CB explanations were the most effective type of {$\mathcal{E}_{err}$} that helps users make informed decisions on a failed plan. Therefore, in this section we introduce an automated explanation generation system that generalizes CB natural language explanations over the failure scenarios and failure types enumerated in Table~\ref{tab:Failures}.  

\subsection{Neural Translation Model Overview }
We adapt a popular encoder-decoder network \cite{bahdanau2015neural,bastings2018annotated} utilized by \cite{ehsan2019automated} to train a model that can generate CB explanations from a set of features obtained from an agent's state.
The set of features, $U$, is comprised of environment features $X$, raw features $N$ and the desired object of interest $o$. As seen in Figure \ref{fig:model-arch}, the network's input features to the encoder include only the environment features. The decoder uses the output of the encoder, H, appended with the raw features, $N$, and the object of interest, $o$, to generate a sequence of target words $Y= \{y_{1},y_{2}...y_{m}\}$, where $y_{i}$ is a single word and $Y$ is the CB explanation.

The encoder and decoder are comprised of Gated Recurrent Units (GRU). The encoder processes the input semantic feature set $X = \{x_{1},x_{2}...x_{n}\}$, and produces a set of hidden states $H = \{h_{1},h_{2}...h_{n}\}$, where a hidden state $h_{i} = GRU(x_{i},h_{i-1})$. In other words, each hidden state is derived from the previous hidden state $h_{i-1}$ as well as the current input semantic feature embedding $x_{i}$. The decoder's input, $s_{0}$, is the encoder's output vector concatenated with the raw feature set, $N$. The decoder then generates hidden states, where a single hidden $s_{i} = GRU(s_{i-1},y_{i-1},c_{i})$.  In this case, each hidden state $s_{i}$ is derived from the previous predicted word $y_{i-1}$, previous hidden state $s_{i-1}$ and a context vector $c_{i}$. The context vector represents a weighted attention vector that allows the model to dynamically focus on features from the decoder's previous hidden state,$s_{i-1}$, and the decoder's input vector,$s_{0}$, for producing the current hidden state $s_{i}$. To select an output word $y_{i}$, we apply a softmax function to $s_{i}$ to obtain a probability distribution over all possible output words and choose the most probable word.

\subsection{Feature Set}
Recall from above that the agent's state is defined as $S = S_{e} \cup S_{l} \cup S_{i} \cup S_{k}$. We utilize the agent's state representations to define the model's feature set $U$. Instead of including the names of all entities $S_{e}$ from the environment, we include only entities that are present at the agent's final location, denoted as $Obj_{G}$.
Additionally, instead of including the agent's absolute position, we include it's position relative to the goal location, denoted as $Rel_{a-Goal}$. Similarly, we include the \textit{minimum} relative distance between objects in $Obj_{G}$ and the desired object $o \in S_{o}$, as $Rel_{o-Obj_{G}}$, and the relative distance between the desired object $o$ and the agent as $Rel_{a-o}$. We also include the agent's angular $v_{ang}$ and linear $v_{lin}$ velocity as well all task states in $S_{k}$. Recall that $S_{k}$ is comprised of $\{k_{grasp},k_{findgrasp},k_{move},k_{pick},k_{detect},k_{seg}\}$ and describes the status of the agent's actions in $A$. Furthermore, we define $o_{p}$ which represents whether $o \in Obj_{G}$ is true or false. Therefore, our environment and raw feature sets are defined as follows: $X = \{Obj_{G}\}$, $N = \{Rel_{a-Goal},Rel_{o-Obj_{G}},Rel_{a-o}, v_{ang}, v_{lin}, S_{k},o_{p}\}$.

\subsection{Data Collection \& Processing}
For our data set, we collected 54 videos, representing each failure scenario from Table \ref{tab:Failures}. For each video, we sampled the collected data at 1 Hz to obtain a holistic representation of the agent's state when executing a plan $\pi$. In addition to annotating each failure state in $\pi$ for each video, we annotated all successful states leading up to the failure state. Given our task objective, some examples of successful states included, ``robot moving to the dining table",``robot has segmented objects in the scene," and ``robot has found grasps for the desired object". To differentiate these annotations from $\mathcal{E}_{err}$, we denote explanations of successful actions as $\mathcal{E}_{corr}$. In this work, $\mathcal{E}_{corr}$ explanations were only used in model training and were not a focus in the experimental evaluation above. In regard to task states in $S_{k}$, we assumed that any value in a given task state stays valid until a subsequent change overrides the current state. Additionally, any empty features were assigned an `Empty' token that the model disregarded via masking.

\subsection{Model Training}
Our model is trained using a two-step grouped leave one out cross validation (LOOCV) with 10 folds. Our LOOCV consists of leaving out an entire scenario of data (25-30 data points) from each possible scenario in Table \ref{tab:Failures}. The first LOOCV is utilized to populate the training set, while the second is used to populate the validation set. Based on the validation loss, on average, our model finishes training in 180 epochs.
We train with a batch size of 20. Our GRU cells in the encoder have a hidden vector size of 20 and the GRU cells in the decoder have a hidden vector size of 49 which accounts for additional raw features, $N$ and the embedding size of $o$. We train our model using a Cross Entropy classification loss optimized via Adam with a learning rate of 0.0001.

\begin{figure}[t!]
\centering
  \includegraphics[width=0.9\columnwidth]{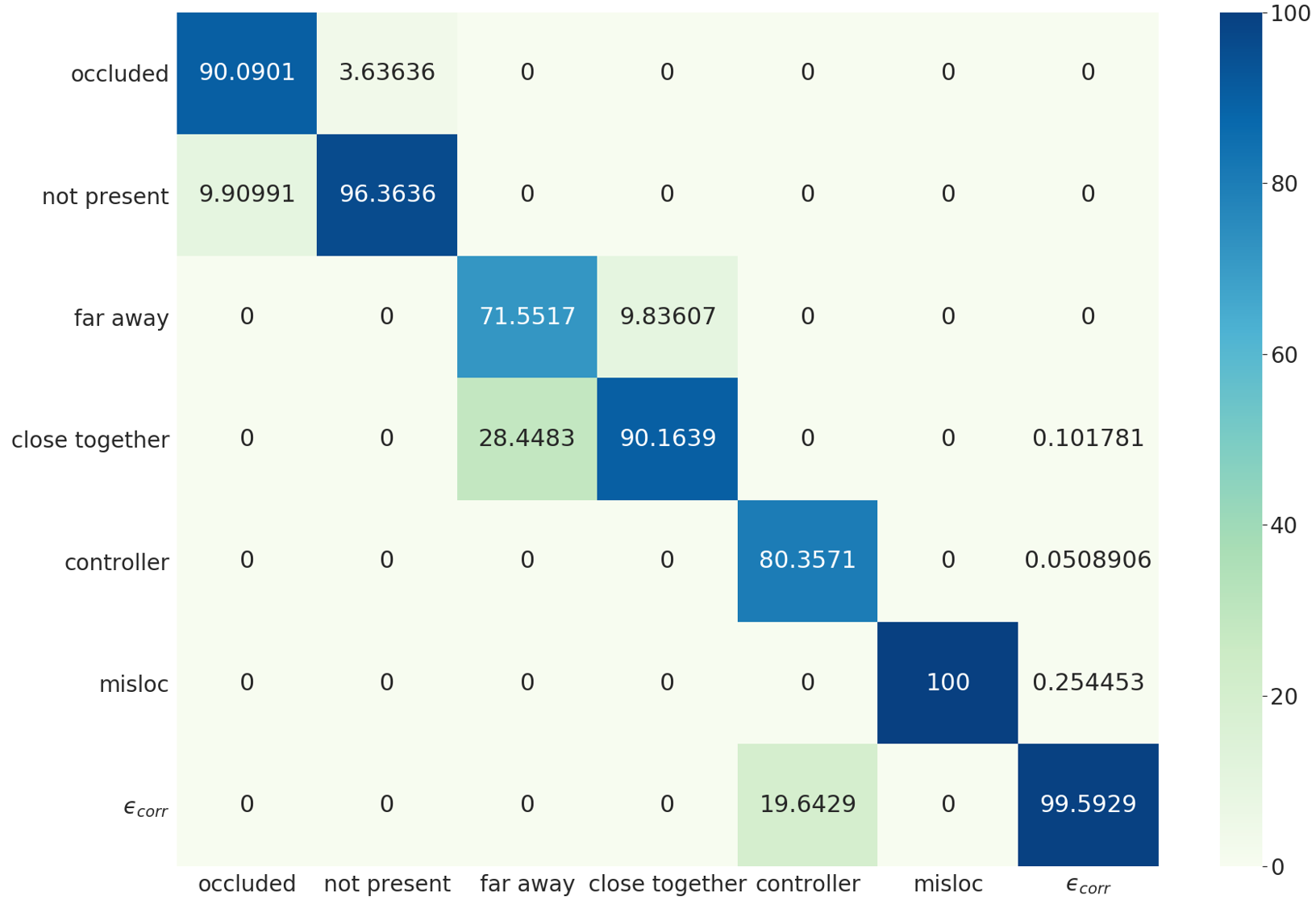}
  \caption{Confusion matrix analysis of our model's performance where the first six columns represent $\mathcal{E}_{err}$ explanations and the last column represents  $\mathcal{E}_{corr}$ explanations.}~\label{fig:CM}
\end{figure}

\subsection{Quantitative Model Evaluation}
Figure \ref{fig:CM} presents the performance of our model across both the six failure explanations $\mathcal{E}_{err}$ presented in Table \ref{tab:Exps} as well as non-error explanations, $\mathcal{E}_{corr}$. In our evaluation, a predicted phrase is only marked correct if it identically matches its target phrase.

On average, our model is able to generalize failure scenarios with a 89.7\% accuracy. We observe that for each failure scenario, the model has a much larger true positive percentage than false positive or false negative percentage. Furthermore, we see that for each failure scenario under the failure types `detection' and `motion-planning' from Table \ref{tab:Failures}, the false positives are within the same failure type. For example a `not present' explanation is only wrongly generated as an `occluded' explanation, both of which are a `detection' failure type. Similarly, `far away' is only wrongly generated as a `close together' explanation and vice versa; both of these failure scenarios fall under the `motion-planning' failure type. However, the failure scenario `controller', under the `navigation' failure type, does not follow this same pattern. Although the `controller' error does not get wrongly predicted as any of the other failure scenarios, it is incorrectly predicted as a 
correct navigation explanation 19.6 percent of the time. Additionally, while we do not analyze the false positives within differing $\mathcal{E}_{corr}$, we do observe that  the non-error explanations are rarely confused with any of the $\mathcal{E}_{err}$ explanations.  

\section{Conclusion}
In this work, we have introduced a new format of explanations, context-based explanations, that is meaningful to a non-expert in not only understanding the failed action in a plan, but also in selecting a recovery solution for a failed action. To validate our context-based explanations, we evaluated it in the domain of a pick-and-place manipulation robot task and investigated users' accuracy in failed action identification, correct recovery identification as well as self-reported ratings of confidence and difficulty in selecting an answer. The results from our user study show that for explanations to be effective in aiding non-expert users to select accurate failure recoveries, the explanations need to include environmental context. The CB explanations allowed users to more effectively select the cause of failure, and the correct failure recovery technique than those who received no explanations.
Additionally, we have we have adapted an existing neural translation model from \cite{ehsan2019automated} to develop automated, CB explanations. The accuracy scores from the confusion matrix show our model's ability to generalize and generate these CB explanations for varied failure scenarios.

This work is motivated to aid non-expert users understand failures that an AI agent may encounter while executing a plan to in turn provide effective failure recovery solutions. Although it includes important contributions, there are limitations that should be addressed in future work. First, while the CB explanations are significantly more useful for assisting in failure recovery than AB or no explanations, they still are not guaranteed to be useful to \textit{all} non-expert users. Therefore future work entails being able to tailor explanations to individual users using reinforcement learning techniques similar to those found in recommender systems \cite{wang2018reinforcement}. Furthermore, our automated explanation generation model can so far generalize over varying failure scenarios. However, a next progression would be to also extend the current model to generalize over varying environments and varying tasks.

\section{Acknowledgments}
This material is based upon work supported by the NSF Graduate
Research Fellowship under Grant No. DGE-1650044.

\bibliographystyle{aaai}
\bibliography{references}

\end{document}